\documentclass[conference]{IEEEtran}
\IEEEoverridecommandlockouts
\usepackage{cite}
\usepackage{amsmath,amssymb,amsfonts}
\usepackage{algorithmic}
\usepackage{graphicx}
\usepackage{textcomp}
\usepackage{xcolor}
\usepackage{balance}
\usepackage{booktabs}

\def\BibTeX{{\rm B\kern-.05em{\sc i\kern-.025em b}\kern-.08em
    T\kern-.1667em\lower.7ex\hbox{E}\kern-.125emX}}
\begin{document}

\title{LSS3D: Learnable Spatial Shifting for Consistent and High-Quality 3D Generation from Single-Image}


\author{
\IEEEauthorblockN{Zhuojiang Cai\IEEEauthorrefmark{3}\IEEEauthorrefmark{1} \quad Yiheng Zhang\IEEEauthorrefmark{4}\IEEEauthorrefmark{1} \quad Meitong Guo\IEEEauthorrefmark{2} \quad Mingdao Wang\IEEEauthorrefmark{2} \quad Yuwang Wang\IEEEauthorrefmark{2}\IEEEauthorrefmark{5}}\\
\IEEEauthorblockA{\IEEEauthorrefmark{2}Tsinghua University, Beijing, China \quad \IEEEauthorrefmark{3}Beihang University, Beijing, China}
\IEEEauthorblockA{\IEEEauthorrefmark{4}National University of Singapore, Singapore}
\IEEEauthorblockA{\IEEEauthorrefmark{1}Equal Contribution}
\IEEEauthorblockA{\IEEEauthorrefmark{5}Corresponding Author \quad wang-yuwang@mail.tsinghua.edu.cn}
}


\maketitle

\begin{abstract}
Recently, multi-view diffusion-based 3D generation methods have gained significant attention. However, these methods often suffer from shape and texture misalignment across generated multi-view images, leading to low-quality 3D generation results, such as incomplete geometric details and textural ghosting. Some methods are mainly optimized for the frontal perspective and exhibit poor robustness to oblique perspective inputs. In this paper, to tackle the above challenges, we propose a high-quality image-to-3D approach, named LSS3D, with learnable spatial shifting to explicitly and effectively handle the multi-view inconsistencies and non-frontal input view.
Specifically, we assign learnable spatial shifting parameters to each view, and adjust each view towards a spatially consistent target, guided by the reconstructed mesh, resulting in high-quality 3D generation with more complete geometric details and clean textures. Besides, we include the input view as an extra constraint for the optimization, further enhancing robustness to non-frontal input angles, especially for elevated viewpoint inputs. We also provide a comprehensive quantitative evaluation pipeline that can contribute to the community in performance comparisons. Extensive experiments demonstrate that our method consistently achieves leading results in both geometric and texture evaluation metrics across more flexible input viewpoints.
\end{abstract}


\section{Introduction}
\label{sec:intro}

Generating 3D content from a single image is of significant value for applications in fields such as virtual reality, gaming, and robotics~\cite{shi_zero123_2023, long_wonder3d_2024, hong_lrm_2023, xu_instantmesh_2024}. However, this task is inherently ill-posed, as it requires not only the reconstruction of visible parts of the image but also the consistent generation of 3D content in unseen regions, making it a challenging problem.
Recently, with the success of 2D image diffusion models~\cite{rombach_high-resolution_2022} in producing high-quality and generalizable results, researchers have extended these models to 3D generation tasks, achieving impressive results in both text-to-3D and image-to-3D applications. However, the Score Distillation Sampling (SDS) based methods~\cite{poole_dreamfusion_2022, wang_prolificdreamer_2024} require long optimization times and may produce unsmooth and inconsistent geometry and textures. Some end-to-end approaches use fast feed-forward models \cite{hong_lrm_2023, xu_instantmesh_2024, xu_grm_2024, boss_sf3d_2024}, which have been trained to generate 3D content from single images, but they are also constrained by computational costs, resulting in limited resolution and difficulty in producing high-quality geometry and textures. Although some recent approaches \cite{hunyuan3d22025tencent, xiang2024structured} have improved generation quality, they struggle to maintain consistency with the input image. This limitation becomes particularly evident when handling intricate facial details, text, and photorealistic elements, where distortions and inconsistencies often emerge.


\begin{figure}[]
 \centering
 \includegraphics[width=\linewidth]{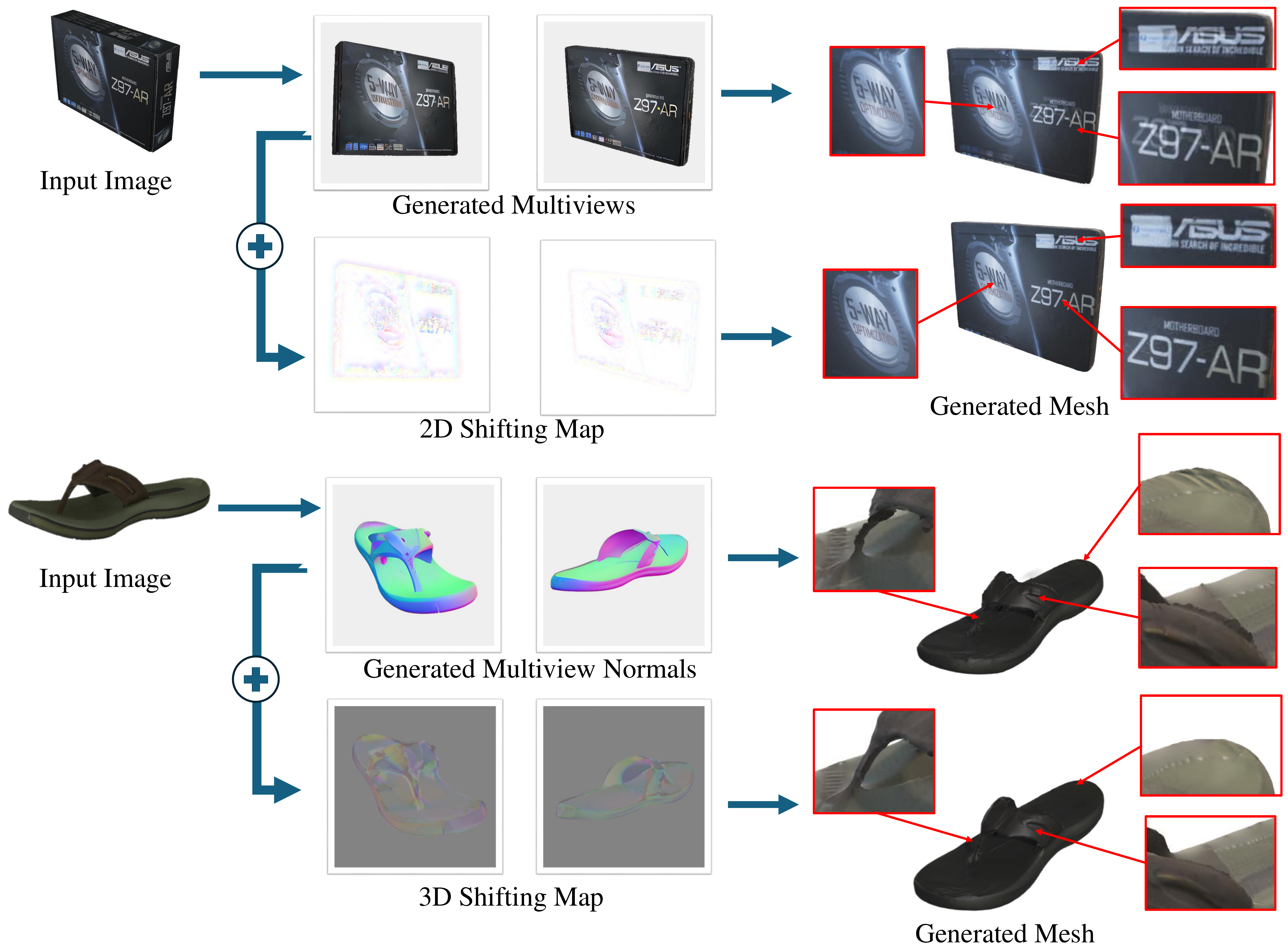}
 \vspace{-7mm}
 \caption{The results with the shift operation (2D shifting on texture and 3D
  shifting on normal maps) exhibit clearer textures and more accurate geometry. The two cases are from GSO dataset\cite{downs_google_2022} .}
  \vspace{-5mm}
 \label{fig:2d3dshift}
\end{figure}

Another type of approach, multi-view diffusion-based methods, has gained significant traction and shows the potential to create high-resolution meshes while maintaining strong consistency with the input image~\cite{wu_unique3d_2024}. 
These methods fine-tune image diffusion models \cite{rombach_high-resolution_2022} on large-scale 3D datasets \cite{deitke_objaverse_2023, deitke_objaverse-xl_2024} to generate multi-view images from a single-view input, then use the generated pseudo multi-views to reconstruct the 3D content \cite{shi_zero123_2023, liu_one-2-3-45_2024, long_wonder3d_2024, wu_unique3d_2024}.  These methods leverage the generalization ability of well-trained 2D image diffusion models and are faster than SDS-based methods, resulting in a significant amount of work. Nonetheless, these methods face challenges with multiview inconsistency in the generated pseudo multiview images, leading to suboptimal quality. These inconsistencies often manifest as misalignment in 3D space. For example, as shown in Figure~\ref{fig:2d3dshift} top row of each case, misalignment textures can generate ghosting appearance when applying texture mapping to the mesh, and the inconsistent normals lead to inaccurate geometric structures.

To address spatial inconsistencies in generated multi-view images, we propose LSS3D, a novel image-to-3D method designed for high-quality 3D reconstruction. Our approach introduces learnable shifting maps for each generated view, enabling explicit spatial adjustments while simultaneously optimizing the mesh. These shifting maps are refined to align the adjusted views with those rendered from the evolving mesh, which serves as a spatially consistent reference. Instead of fitting the mesh to the raw, inconsistent views, we optimize it to match the adjusted ones, ensuring a more stable and accurate reconstruction. To prevent trivial solutions, i.e., both the mesh and spatial shifting converge to a low-quality state, we incorporate a coarse mesh initialization, enforce smoothness constraints to limit excessive shifts, and include the input view in the optimization process. Our approach improves both geometry and texture quality by first optimizing 2D and 3D shifting maps of the normal maps for each view to achieve a consistent shape. Specifically, 2D shifting indicate spatial deformation and 3D shifting introduce normal value adjustment. We then further refine the 2D shifting maps to correct minor texture misalignments, ensuring high-fidelity reconstruction with improved detail preservation.

Although some recent works~\cite{long_wonder3d_2024, wu_unique3d_2024, boss_sf3d_2024} present quantitative metrics in terms of both geometry and texture, the specific evaluation data and details of the comparisons remain unclear, and there is significant variation in evaluation results across different works. 
To address this issue, we provide a detailed and automated evaluation pipeline and select all the single object cases (927 cases) from GSO dataset\cite{downs_google_2022}.

In summary, our main contributions are as follows:

\begin{enumerate}
    \item We propose LSS3D, a novel image-to-3D approach that efficiently and explicitly addresses multi-view inconsistencies and input-view camera pose estimation, enabling high-quality mesh generation from single images.
    \item We introduce two shifting strategies to mitigate inconsistencies in pseudo multi-view images: 2D shifting to reduce ghosting while preserving more details, and 3D shifting to ensure smoother surfaces.
    \item Extensive experiments demonstrate the effectiveness of our method in enhancing 3D generation quality, making it a robust framework with high generalization, fidelity, and fine-grained detail representation.
\end{enumerate}



\section{Method}

\begin{figure*}[]
 \centering
 \includegraphics[width=0.85\linewidth]{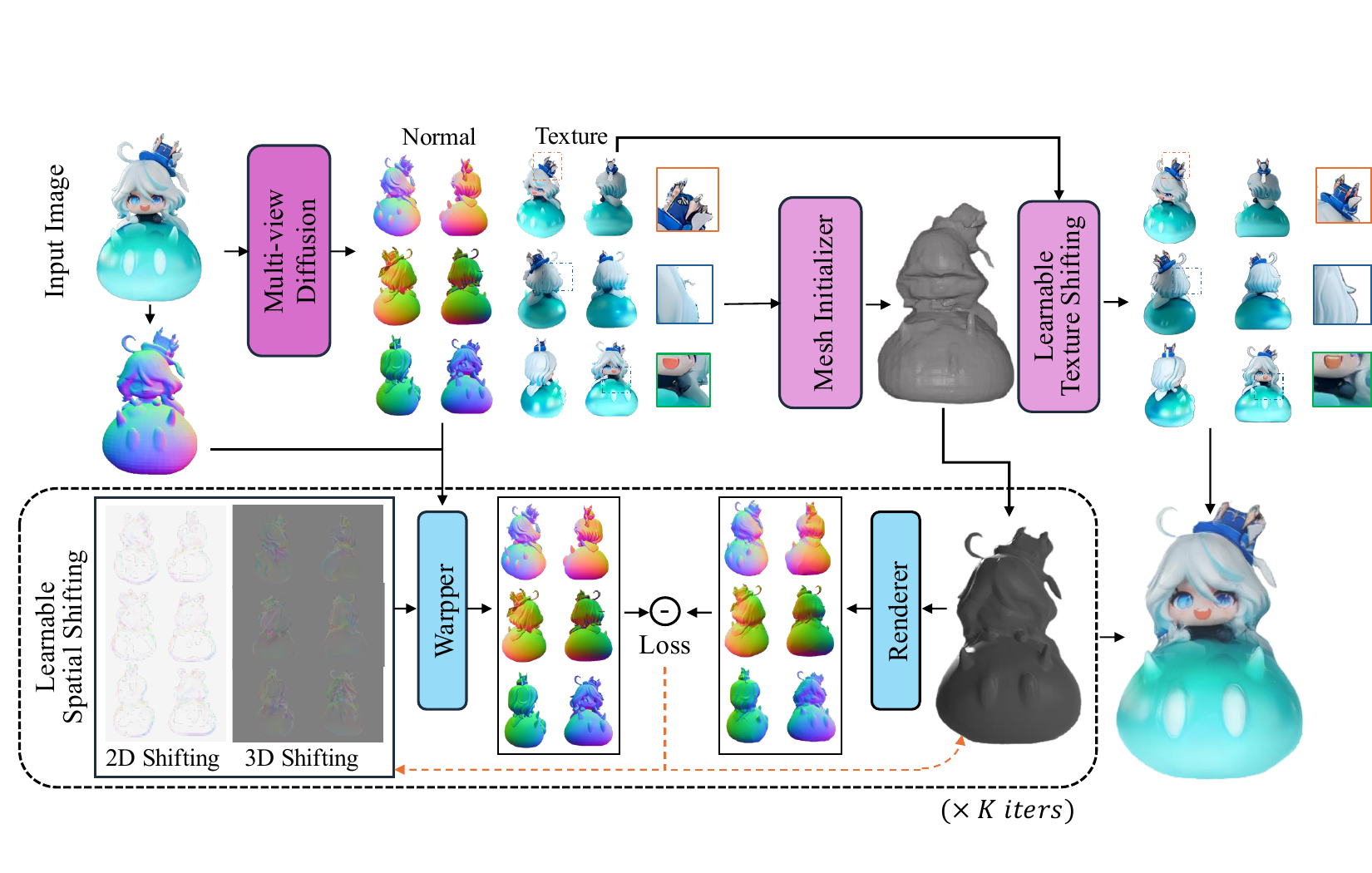}
 \vspace{-3mm}
 \caption{\textbf{Overview of LSS3D.} Given an image of an object, our method first employs a multi-view diffusion model to generate six-view images and normal maps, which are then used to quickly reconstruct a coarse mesh. Next, the normal maps are used to iteratively optimize the mesh, with the normal maps passing through a learnable spatial shifting wrapper. Both the shifting maps and the mesh are optimized, which is completed quickly (\(\sim\)10 s). Finally, multi-view images, adjusted with a texture shifting wrapper, are projected onto the mesh.}
  \vspace{-3mm}
 \label{fig:method}
\end{figure*}

In this section, we introduce LSS3D, a multi-view diffusion model-based framework for efficiently generating textured meshes from single images. Our method first employs a multi-view diffusion model to generate pseudo multi-view images and their corresponding normal maps, along with a mesh initializer that constructs a coarse mesh. The camera pose of the input image is then estimated to facilitate its use in subsequent optimization and projection (Sec. \ref{sec:pose_estimation}). Next, the coarse mesh is iteratively optimized using the multi-view normal maps, aided by a spatial shifting module that mitigates 3D inconsistencies across views, enabling seamless integration of details from all perspectives (Sec. \ref{sec:spatial_shifting}). The geometry optimization process takes approximately 10 seconds and produces a high-fidelity, consistent mesh (Sec. \ref{sec:geometry_optimization}). Finally, the multi-view images are used to optimize a texture shifting module, which helps prevent artifacts and inconsistencies during projection. The adjusted images are then projected onto the mesh (Sec. \ref{sec:texture_projection}). An overview of our framework is shown in Figure \ref{fig:method}.

\subsection{Input-View Camera Pose Estimation}
\label{sec:pose_estimation}

Since users typically want the generated mesh to be as consistent as possible with the input image, it is important to incorporate the input image throughout the entire generation process. Zero123++ implicitly corrects the elevation origin to accommodate input images with an elevated view, and thus the elevation of the input image needs to be estimated in order to integrate it into the subsequent mesh optimization process. We estimate it by maximizing the similarity \( s^{(elev)} \) using a coarse-to-fine search over renderings with different elevations: 
\begin{equation}
    s^{(elev)} = \sum_{x = 1}^{w} \sum_{y = 1}^{h} (\hat{N}^{(elev)}(x, y) \cdot N_0^{pseudo}(x, y)),
\end{equation}
where \( \hat{N}^{(elev)} \) is the rendered normal map from the mesh at elevation \( elev \) and azimuth \( 0 \), and \( N_0^{pseudo} \) is the pseudo normal map for the input view. \( w \) and \( h \) represent the width and height of the image, respectively. The final elevation is estimated after several iterations within the initial range \( [b_0, t_0] \), with each iteration sampling \( n_s \) views linearly. When the maximum similarity elevation \( e_i \) is found in the \( i \)-th iteration, the search range for the next iteration becomes \( [e_i - \epsilon_i, e_i + \epsilon_i] \), where \( \epsilon_i = (t_i-b_i)/n_s \) is the sampling interval.

\subsection{Spatial Shifting}
\label{sec:spatial_shifting}

The multi-view images generated by diffusion models often exhibit inconsistencies across views, leading to surface roughness, texture ghosting, and detail loss when used as targets for mesh optimization. If these multi-view images and normal maps could be enhanced for consistency through some form of warping, better reconstruction quality may be achieved. Therefore, we introduce a learnable spatial shifting mechanism that optimizes alongside the mesh to adapt the normal maps to the current mesh.

The inconsistencies among pseudo multi-view images can be considered in two ways: pixel misalignment between 2D images and discrepancies in the 3D orientation of the same normal vector across normal maps. To address these issues, we propose two types of spatial shifting: 2D and 3D shifting, which improve the consistencies of multi-view images for better mesh reconstruction.

\noindent\textbf{2D Shifting.} Similar to optical flow, we use a shifting map representing per-pixel 2D displacement vectors to adjust the multi-view images and normal maps. 2D shifting effectively corrects pixel misalignment across different views in pseudo multi-view images and normal maps, reducing texture ghosting and detail loss. Given a 2D shifting map \( S_{2D} \in \mathbb{R}^{H\times W\times 2}\), the warping process can be expressed as:
\begin{equation}
    \mathcal{I}_{2D}^{warp} = \text{grid\_sample}(\mathcal{I}^{pseudo}, G_o + S_{2D}) ,
\end{equation}

\noindent where \( \mathcal{I}^{pseudo}, \mathcal{I}_{2D}^{warp} \in \mathbb{R}^{H\times W\times 3}\) are the images or normal maps before and after shifting, \( G_o \) is the coordinate grid of the image, and \( \text{grid\_sample}(\cdot, \cdot) \) is a differentiable grid sampling function.

\noindent\textbf{3D Shifting.} During shape optimization, inconsistencies in the 3D orientation of the same normal vector across different view normal maps may lead to suboptimal geometry, such as rough surfaces. To mitigate this issue, we introduce a 3D shifting map representing per-pixel 3D normal vector displacements to adjust the multi-view normal maps. Given a 3D shifting map \( S_{3D} \in \mathbb{R}^{H\times W\times 3}\), the warping process can be expressed as:
\begin{equation}
    \mathcal{I}_{3D}^{warp} = \mathcal{I}^{pseudo} + S_{3D}.
\end{equation}

During the initialization of the shifting map, we use only 2D shifting to extract 3D consistency information from the initial mesh. As the shifting map and mesh are jointly optimized, both 2D and 3D shifting are employed. During texture shifting optimization, 2D shifting is used to align the input view with textures from other views.

\subsection{Geometry Optimization Process}
\label{sec:geometry_optimization}

Given an image of an object, LSS3D first employs the multi-view diffusion model Zero123++ \cite{shi_zero123_2023} to generate six-view images and their corresponding normal maps. An initializer is then applied to create a coarse mesh from these six views. The coarse mesh facilitates the subsequent mesh optimization and provides a reference for the elevation estimation of the input view.

Then, our method directly optimizes the mesh, because mesh-based optimization \cite{wu_unique3d_2024} requires computation proportional to the square of the resolution, enabling faster generation of high-quality content. In contrast, field-based reconstruction is computationally expensive, as its complexity scales with the cube of the resolution, making it challenging to achieve high-resolution results. During the optimization process, the mesh can be differentiably rendered from multiple views, and the loss and gradients are computed with respect to the target multi-view images. The optimization minimizes the loss through vertex movement, edge collapse, edge splitting, and edge flipping.

\noindent\textbf{Loss Function.} To optimize both the mesh and the shifting maps, the loss function first includes the L2 loss for each view between the rendered and warped normal maps:
\begin{equation}
    \mathcal{L}_{L2} = \sum_i{|| \hat{N}_i -  N_i^{warp} ||},
\end{equation}
and the mask loss to constrain the mesh alignment with the edges of the normal maps:
\begin{equation}
    \mathcal{L}_{mask} = \sum_i{|| \hat{M}_i -  M_i^{warp} ||},
\end{equation}
where \( \hat{N}_i \) and \( N_i^{warp} \) represent the rendered and warped normal maps under view \( i \), respectively, and \( \hat{M}_i \) and \( M_i^{warp} \) are their corresponding masks. Additionally, a smoothness loss is applied to enforce smoothness in both the shifting maps and the warped normal maps. The \( k \)-th order smoothness loss is defined as:
\begin{equation}
    {\mathcal{L}}_{smooth(k)}(X)=\sum_i{\sum\limits_{m\in\{x,y\}}w_{edge}\left|\frac{\partial^k X_i}{\partial m^{k}}\right|},
\end{equation}
where \( X_i \) represents the maps to be smoothed under view \( i \), \(w_{edge}=\exp\left( - \lambda_{edge}\sum\limits_{c}\left|\frac{\partial N_i^{pseudo}}{\partial m}\right|\right)\) is the edge weight to reduce smoothing strength along edges, and \( c \) represents the normal channels. We apply second-order smoothness on the 2D and 3D shifting maps, and first-order and second-order smoothness on the warped normal maps.

Finally, the loss function for mesh optimization with spatial shifting is given by:
\begin{equation}
\begin{split}
    {\mathcal{L}}={\mathcal{L}}_{L2}+\lambda_{1}{\mathcal{L}}_{mask}+\lambda_{2}{\mathcal{L}}_{smooth},
\end{split}
\end{equation}
where \(\lambda_{1}\) and \(\lambda_{2}\) are the weights for each of the respective loss terms, respectively.

\noindent\textbf{Optimization Details.} To utilize the 3D information of the initialized mesh, the optimization process first focuses solely on optimizing the 2D shifting while keeping the mesh fixed. This step serves as the initialization phase for the shifting map. Once the 2D shifting map has captured the initial image displacements, the joint optimization of both the shifting map and the mesh begins, incorporating both 2D and 3D shifting maps.

The supervision for shifting maps is performed at multiple scales. In addition to the original resolution, the shifting maps are downsampled to 1/2 and 1/4 of the size and used to warp the images. Losses are calculated at these three scales to helps learn shifting at different step sizes, similar to the approach used in optical flow estimation. Additionally, the 3D shifting is reset every \(K\) iterations to capture the finer details of the original multi-view normal maps, preventing excessive smoothing and deformation in the generated mesh. The 2D shifting is maintained as a cumulative process throughout the optimization.

Both the six-view and input view normal maps undergo spatial shifting warping and are involved in the optimization process. However, a smaller learning rate is applied to the shifting maps of the input view, as we aim to place higher confidence in the input image and encourage the other views to align with it. The mesh optimization process requires several hundred iterations, taking around 10 seconds to complete on a single RTX 4090.

\subsection{Texture Projection with Shifting}
\label{sec:texture_projection}

After completing the geometry optimization, our method projects the multi-view images onto the mesh to generate a textured mesh. Similar to the geometry optimization step, the multi-view images first undergo 2D shifting to obtain warped images. The 2D shifting maps are derived from the results of geometry optimization and are further refined by optimizing the alignment between the warped multi-view images and the reprojected input view image in other views, ensuring better consistency and reducing artifacts.

Next, the warped multi-view images are projected onto the vertices of the mesh using cosine-weighted blending. For a vertex \( v \), its projected color \( Col(v) \) is computed as:
\begin{equation}
    Col(v)=\frac{\sum_i{V(v,i)(n^{view}_i\cdot n_v)^2Col_\mathcal{I}(v,\mathcal{I}^{warp}_i)}}{\sum_i{V(v,i)}}
\end{equation}
where \( V(v,i) \in \{0, 1\} \) indicates the visibility of vertex \( v \) from view \( i \), \( n^{view}_i \) is the view direction of view \( i \), \( n_v \) is the normal of vertex \( v \), and \( Col_\mathcal{I}(v, \mathcal{I}^{warp}_i) \) is the color of vertex \( v \) in the warped image \( \mathcal{I}_i^{warp} \). The weight \( (n^{view}_i \cdot n_v)^2 \) is determined by the square of the cosine of the angle between the view direction and the vertex normal. This cosine weighting helps ensure smooth blending of the texture at the intersection between views.

With texture shifting and projection, our method generates clean, textured meshes, ensuring that the final result preserves both geometric and visual fidelity.
\section{Experiments}

\begin{table}[t]
    \caption{Quantitative comparison for visual and geometry quality. We report the metrics on the built dataset from GSO.}
    \vspace{-3mm}
  \label{tab:quant_full}
  \centering
  \small
  \resizebox{\linewidth}{!}{ 
  \begin{tabular}{lccccccc}
  \toprule
  \textbf{Method} & \textbf{PSNR$\uparrow$} & \textbf{SSIM$\uparrow$}  & \textbf{LPIPS$\downarrow$} & \textbf{Clip-Sim$\uparrow$} & \textbf{CD$\downarrow$} & \textbf{F-Score (0.1)$\uparrow$} \\
  \midrule  
  InstantMesh \cite{xu_instantmesh_2024}     
  & 18.3521          & 0.8005          & 0.1867          & 0.9045             & 0.05989                 & 0.8138\\
  TripoSR \cite{tochilkin_triposr_2024}
   & 18.8082          & 0.7977          & 0.1951          & 0.9198             & 0.0581                & 0.8105\\
  Unique3D \cite{wu_unique3d_2024}        
  & 18.4918          & 0.7955          & 0.1958          & 0.8914             & 0.0770                & 0.7061\\
  Wonder3D++ \cite{long_wonder3d_2024}        
  & 18.6989          & 0.8047          & 0.1779          & 0.8911             & 0.0603                 & 0.8112\\
  SF3D \cite{boss_sf3d_2024}           
  & 18.6780          & 0.8105          & 0.1860        & 0.9169             & 0.0557                 & 0.8244\\
  
  \midrule
  LSS3D (Ours)         & \textbf{20.7075}         & \textbf{0.8492}          & \textbf{0.1448}          & \textbf{0.9371}             & \textbf{0.0319}                 & \textbf{0.9387}\\
  \bottomrule
  \end{tabular}
  }
  \vspace{-2mm}
\end{table}

\subsection{Experimental Setting}

\textbf{Pretrained Models.} For the multi-view diffusion model, we use Zero123++ \cite{shi_zero123_2023}, as it demonstrates more robust performance when handling elevated viewpoint input images compared to multi-view diffusion models based on orthogonal views \cite{long_wonder3d_2024, wu_unique3d_2024}. To initialize the mesh, we use a fast feed-forward model, InstantMesh \cite{xu_instantmesh_2024}, the effectiveness of which is confirmed through an ablation study.

\noindent\textbf{Datasets.} 
Following previous studies \cite{liu_syncdreamer_2023, long_wonder3d_2024, wu_unique3d_2024}, we adopt the Google Scanned Objects (GSO) dataset \cite{downs_google_2022} for evaluation. The GSO dataset is a large-scale collection of 3D objects, comprising a total of 1,032 cases. To ensure a comprehensive evaluation, we exclude cases with multiple objects, resulting in a final dataset of 927 cases.

\noindent\textbf{Metrics.} We employ PSNR, SSIM \cite{wang_image_2004}, LPIPS \cite{zhang_unreasonable_2018}, and CLIP similarity \cite{radford_learning_2021} to assess the visual quality of rendered multi-view images from these meshes. We use Chamfer Distance (CD) and F-score to evaluate the geometric quality of the generated meshes. We also design new metrics to evaluate the multiview consistency.

\noindent\textbf{Evaluation Pipeline.} For the generated meshes, we automate the rotation and scaling to align them with the ground truth, then normalize them to the range \([-0.5, 0.5]\). We select elevation angles from \([0^\circ, 15^\circ, 30^\circ]\) and eight azimuth angles evenly distributed across a full 360-degree rotation, resulting in a total of 24 views. The images are rendered at a resolution of \(1024\times 1024\).

\subsection{Main Results}

\begin{figure}[htbp]
    \centering
    \includegraphics[width=\columnwidth]{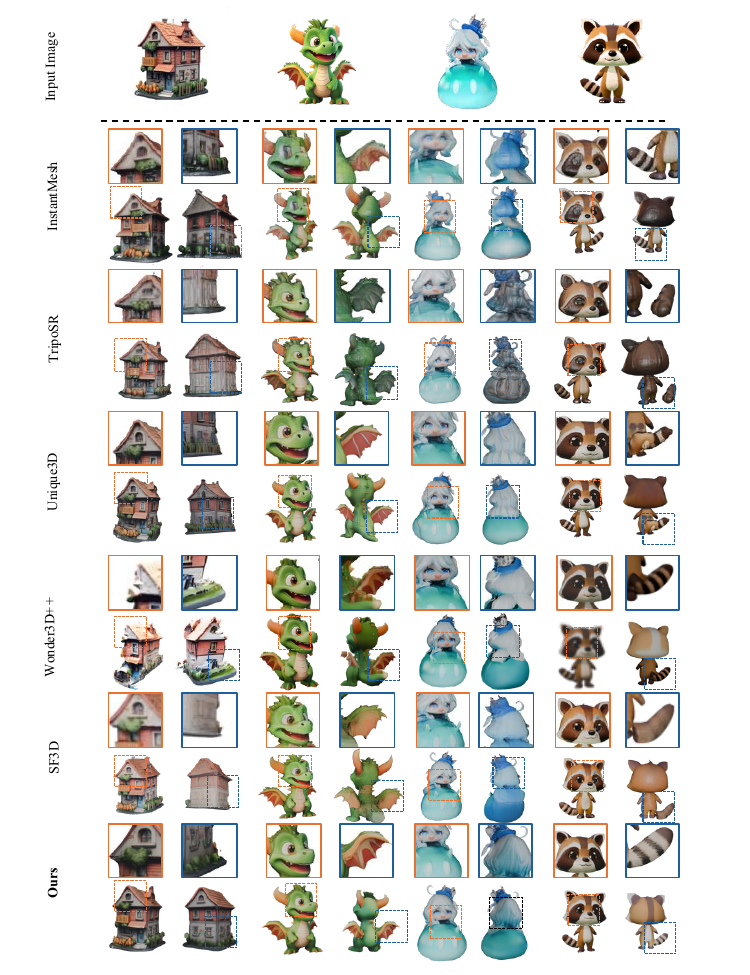}
    \caption{Qualitative comparison. Our approach provides smooth and detailed geometry with clean texture. (Zoom in to see more details.)}
    \label{fig:qualitative}
\end{figure}

\noindent\textbf{Qualitative Comparison.} To highlight the advantages of our approach, we compare the generated results with those of previous works, including InstantMesh \cite{xu_instantmesh_2024}, TripoSR \cite{tochilkin_triposr_2024}, Unique3D \cite{wu_unique3d_2024}, Wonder3D++ \cite{long_wonder3d_2024}, and SF3D \cite{boss_sf3d_2024}, using the same sample input images. As shown in Figure \ref{fig:qualitative}, our approach produces smooth and detailed geometry with clean textures, in contrast to the results from other methods.
For instance, in terms of geometric details such as the hair in Figure \ref{fig:qualitative}(b) and the ears in Figure \ref{fig:qualitative}(c), our generation is more complete. 
Furthermore, thanks to the better alignment through spatial shifting optimization, our method produces clearer textures for the face regions in Figures \ref{fig:qualitative}(a), (b), and (c), as well as for the balcony in Figure \ref{fig:qualitative}(d). 
This comparison demonstrates the superiority of our method in ensuring 3D consistency when optimized using multi-view images.

\noindent\textbf{Quantitative Comparison.} Following the experimental setup outlined earlier, we conduct a quantitative comparison between our method and four other previous state-of-the-art approaches, two multi-view diffusion model-based methods (Wonder3D++ and Unique3D), and three representative feed-forward models (InstantMesh, TripoSR and SF3D), using the whole dataset of the GSO dataset, excluding only multi-object files.
The evaluation metrics include PSNR, SSIM, LPIPS, CLIP similarity, Chamfer Distance (CD), and F-score, covering both texture and shape quality. The results, presented in Table \ref{tab:quant_full}, show that our approach outperforms existing methods in terms of both geometric and texture quality. 

\begin{figure}[htbp]
    \centering
    \includegraphics[width=0.5\textwidth]{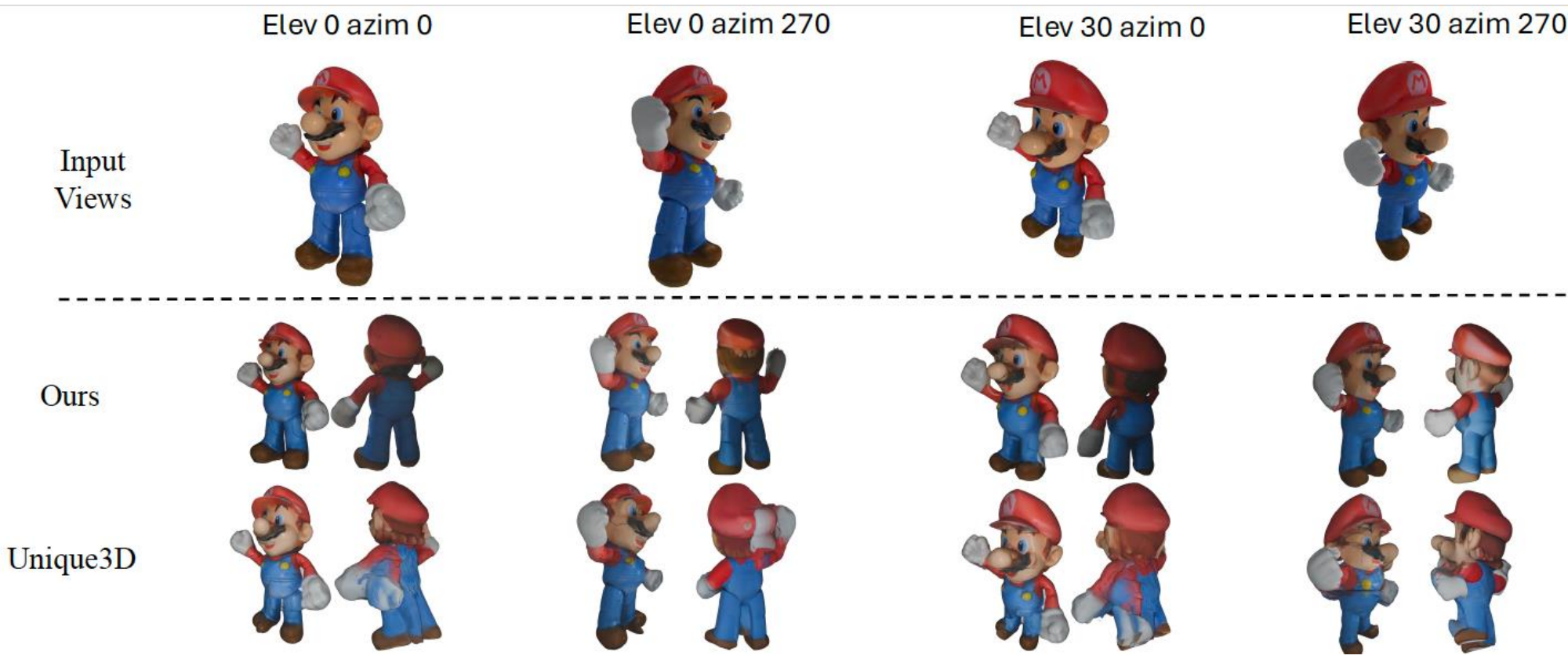}
    \vspace{-3mm}
    \caption{Our method shows higher robustness to input views.}
    \label{fig:inputview}
    \vspace{-3mm}
\end{figure}

\begin{figure}[]
 \centering
 \includegraphics[width=\linewidth]{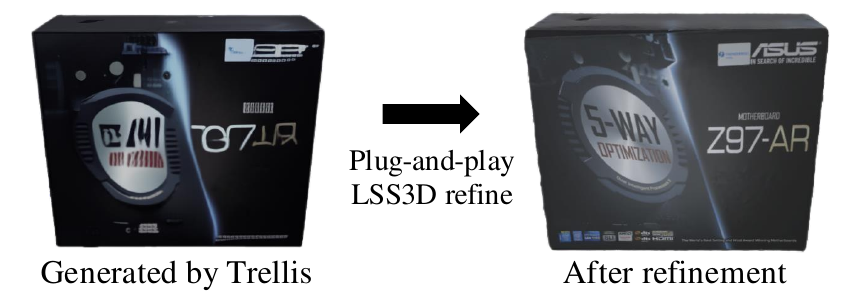}
 \caption{LSS3D plug-and-play refinement of results from other 3D generation methods (e.g., Trellis~\cite{xiang2024structured})}
 \label{fig:plug}
\end{figure}

\begin{figure}[htbp]
    \centering
    \includegraphics[width=\linewidth]{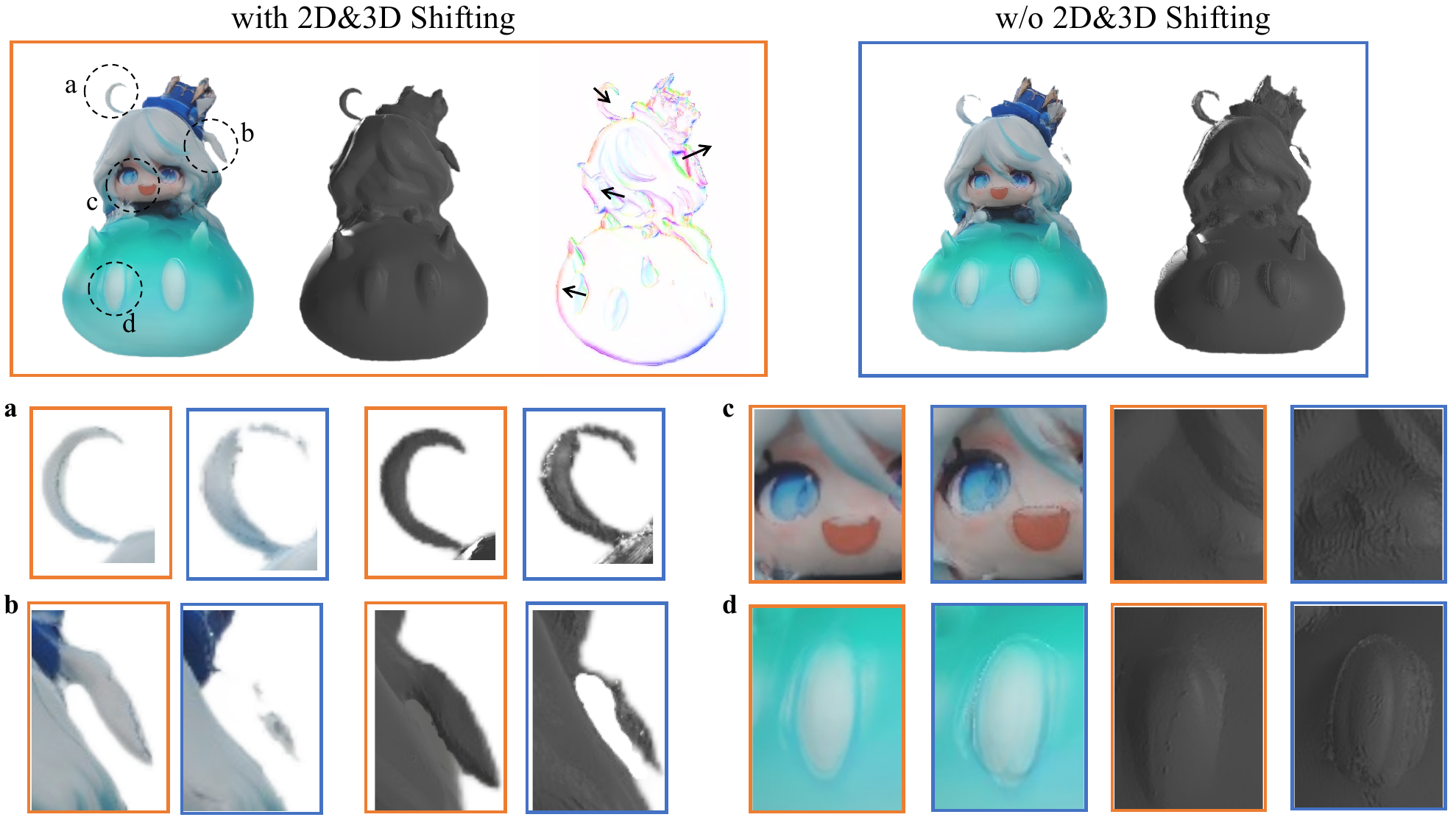}
    \vspace{-5mm}
    \caption{Ablation study. As the generation results shown, with our proposed learnable spatial shifting, the detailed shapes in regions (a) and (b) are reconstructed more completely, and the shape and texture are better aligned, as shown in regions (c) and (d).}
    \vspace{-5mm}
    \label{fig:ablation}
\end{figure}

\noindent\textbf{Input View Robustness Comparison.} 
We also evaluate the robustness of LSS3D to varying input views, which is important for practical applications. As Figure~\ref{fig:inputview} shows, we use 4 different input views to generate 4 meshes, and for each mesh render 2 different views. The results demonstrate that our method generates high-quality 3D model for all input views, significantly surpassing Unique3D.

\noindent\textbf{Plug-and-Play Enhancement.} 
LSS3D can also be used to enhance the quality of results from other 3D generation methods in a plug-and-play manner. For example, the results generated by Trellis~\cite{xiang2024structured} may not perfectly match the input images, leading to ambiguous text and textures. We applied LSS3D's optimization process for refinement, and the results are shown in Figure~\ref{fig:plug}.

\subsection{Ablation Study} 

\begin{table}[t]
    \caption{\textbf{MSE} and \textbf{Flow Error}  with or without spatial shifting.}
    \vspace{-3mm}
  \label{tab:new_metric}
  \centering
  \small
  \resizebox{0.7\linewidth}{!}{ 
  \begin{tabular}{lccccccc}
  \toprule
  \textbf{Settings} & \textbf{MSE$\downarrow$} & \textbf{Flow Error$\downarrow$}  \\
  \midrule 
  w/o shifting & 0.004756 & 16.76\\
  with shifting & \textbf{0.003493} & \textbf{14.60} \\
  \bottomrule
  \end{tabular}
  }
  \vspace{-3mm}
\end{table}

\textbf{Shifting Ablation.} 
We conduct an ablation study to analyze the impact of learnable spatial shifting in our approach. 
Our method, which employs both 2D and 3D shifting, achieves the best overall performance in terms of generation quality.
Further visual comparisons, shown in Figure \ref{fig:ablation}, demonstrate that the reconstructions with 2D and 3D spatial shifting preserve fine details better than those without this technique. 

\noindent \textbf{Inconsistencies Metrics.}
We additionally designed two metrics, MSE and Flow Error, to evaluate the consistency between the multi-view images and the input view, with and without our proposed spatial shifting.
Specifically, we project the images from the two viewpoints adjacent to the input viewpoint onto the input viewpoint and compute the Mean Squared Error (MSE) between the projected images and the input image in the visible regions. Additionally, we estimate the optical flow between the projected images and the input image using the RAFT optical flow estimation method, which we define as the Flow Error. Both metrics reflect the consistency error between the generated images and the input image.
The results (Table \ref{tab:new_metric}) indicate that after shifting, both metrics exhibit a significant decrease (MSE and flow error decrease by 26.5\% and 12.9\%, respectively), indicating that our method indeed improves consistency.

\section{Conclusion}

In this paper, we propose LSS3D, an approach for 3D generation from single images by explicitly optimizing the inconsistency across the images generated by multi-view diffusion models. As inconsistency is a common issue in generated multi-views, LSS3D offers an effective and efficient way to improve 3D generation quality, achieving more complete details and fewer misalignments. Both experimental comparisons to recent leading approaches and ablation studies validate the effectiveness of LSS3D. We believe the basic idea of spatial shifting can be further explored to address region level inconsistencie with implicit diffusion refinement.

\balance
\bibliographystyle{IEEEbib}
\bibliography{main}

\end{document}